%% file: camera_ready.tex
\documentclass[runningheads]{llncs}

\usepackage{eccv}

\newcommand{\ours}{RoMa-\ensuremath{\Omega}}

\newcommand{\RN}[1]{\uppercase\expandafter{\romannumeral #1}}

\usepackage[dvipsnames]{xcolor}

\input{vincents_commands}

\addeditor{vincent}{VL}{0.0, 0.5, 0.0}
\showeditstrue

\usepackage{eccvabbrv}

\usepackage{graphicx}
\usepackage{booktabs}
\usepackage{multirow}

\usepackage[accsupp]{axessibility}  %

\usepackage[pagebackref]{hyperref}
\usepackage{enumitem}
\usepackage{wrapfig}

\usepackage{orcidlink}

\usepackage{array}
\usepackage{graphicx}
\usepackage[percent]{overpic} %
\usepackage[table]{xcolor}

\usepackage[HTML]{xcolor}

\definecolor{romarow}{HTML}{E8F5F0}
\definecolor{romav2row}{HTML}{E3F2FD}
\definecolor{oursrow}{HTML}{FCE4EC}

\begin{document}

\title{\ours: What Feed-Forward 3D Models Know About Image Matching} 

\titlerunning{\ours: What Feed-Forward 3D Models Know About Image Matching}

\authorrunning{D. Nordström et al.}

\institute{$^1$Chalmers University of Technology, Sweden\\ \quad $^2$Mobile Perception Lab, ShanghaiTech University, China\\ \quad $^3$LIGM, Ecole des Ponts, Univ. Gustave Eiffel, CNRS, France
}

\author{
David Nordström$^1$
\and
Xinyue Zhang$^2$
\and
Thibaut Loiseau$^3$ \\
Vincent Lepetit$^3$
\and
Fredrik Kahl$^1$
\\
}
\maketitle

\begin{abstract}
Learned image matching has experienced significant progress in recent years, culminating in robust and accurate matchers such as RoMa, whose robustness is often attributed to its use of frozen DINO features. In a parallel development, feed-forward reconstruction models, such as VGGT, have been trained on ever-growing datasets to accurately regress dense 3D point maps and camera poses. The distinction between matchers and feed-forward reconstruction models has become increasingly blurred with the introduction of matching losses in models such as MASt3R and VGGT-$\Omega$. This raises a natural question: \textit{what do feed-forward 3D models know about image matching?} In this work, we answer this question by analyzing three scenarios: (i) zero-shot matching of patch features, (ii) direct matching of 3D point predictions, and (iii) training a full matcher on top of the learned representations. We find that, despite performing poorly in zero-shot matching, especially in later layers, feed-forward reconstruction models provide strong representations for linear probing and full matching pipelines. We further show that, even without any training, their raw predictions alone enable competitive matching, albeit only under moderate viewpoint changes and modality gaps. Based on these insights, we retrain RoMa v2 by replacing its DINO backbone with VGGT-$\Omega$. Our resulting model, \ours, outperforms state-of-the-art matchers on a wide range of benchmarks, \eg  +8.1 mAA compared to RoMa v2 on WxBS.  We release our code and weights publicly at \href{https://github.com/davnords/RoMa-Omega}{https://github.com/davnords/RoMa-Omega}
\end{abstract}

\input{sec/01_introduction}

\input{sec/02_related_work}
\input{sec/03_method}
\input{sec/04_experiments}
\input{sec/05_limitations}
\input{sec/06_conclusion}
\newpage
\input{sec/acknowledgements}

\bibliographystyle{splncs04}
\bibliography{main}
\input{sec/supplementary}

\end{document}

%% file: vincents_commands.tex
\usepackage{dsfont}
\usepackage{etoolbox}
\usepackage{color}

\newif\ifshowedits

\newcommand{\addeditor}[3]{%
  \definecolor{#1color}{rgb}{#3}
  \expandafter\newcommand\csname #1\endcsname[1]{%
  \ifshowedits
    {\color{#1color} ##1}%
  \else
    {##1}%
  \fi
  }%
  \expandafter\newcommand\csname #1rmk\endcsname[1]{%
  \ifshowedits
    {\color{#1color} {\bf [#2: ##1]}}
  \fi
  }%
  \expandafter\newcommand\csname #1rpl\endcsname[2]{%
  \ifshowedits
    {\color{#1color} ##1 \sout{##2}}
  \else
    {##1}
  \fi
  }%
}

\newcommand{\createtextvar}[1]{
  \expandafter\newcommand\csname #1\endcsname{%
  {\text{#1}}
}%
}

\usepackage[bigfiles]{pdfbase}
\ExplSyntaxOn
\NewDocumentCommand\embedvideo{smm}{
  \group_begin:
  \leavevmode
  \tl_if_exist:cTF{file_\file_mdfive_hash:n{#3}}{
    \tl_set_eq:Nc\video{file_\file_mdfive_hash:n{#3}}
  }{
    \IfFileExists{#3}{}{\GenericError{}{File~`#3'~not~found}{}{}}
    \pbs_pdfobj:nnn{}{fstream}{{}{#3}}
    \pbs_pdfobj:nnn{}{dict}{
      /Type/Filespec/F~(#3)/UF~(#3)
      /EF~<</F~\pbs_pdflastobj:>>
    }
    \tl_set:Nx\video{\pbs_pdflastobj:}
    \tl_gset_eq:cN{file_\file_mdfive_hash:n{#3}}\video
  }
  \pbs_pdfobj:nnn{}{dict}{
    /Type/RichMediaInstance/Subtype/Video
    /Asset~\video
    /Params~<</FlashVars (
      source=#3&
      skin=SkinOverAllNoFullNoCaption.swf&
      skinAutoHide=true&
      skinBackgroundColor=0x5F5F5F&
      skinBackgroundAlpha=0.75
    )>>
  }
  \pbs_pdfobj:nnn{}{dict}{
    /Type/RichMediaConfiguration/Subtype/Video
    /Instances~[\pbs_pdflastobj:]
  }
  \pbs_pdfobj:nnn{}{dict}{
    /Type/RichMediaContent
    /Assets~<<
      /Names~[(#3)~\video]
    >>
    /Configurations~[\pbs_pdflastobj:]
  }
  \tl_set:Nx\rmcontent{\pbs_pdflastobj:}
  \pbs_pdfobj:nnn{}{dict}{
    /Activation~<<
      /Condition/\IfBooleanTF{#1}{PV}{XA}
      /Presentation~<</Style/Embedded>>
    >>
    /Deactivation~<</Condition/PI>>
  }
  \hbox_set:Nn\l_tmpa_box{#2}
  \tl_set:Nx\l_box_wd_tl{\dim_use:N\box_wd:N\l_tmpa_box}
  \tl_set:Nx\l_box_ht_tl{\dim_use:N\box_ht:N\l_tmpa_box}
  \tl_set:Nx\l_box_dp_tl{\dim_use:N\box_dp:N\l_tmpa_box}
  \pbs_pdfxform:nnnnn{1}{1}{}{}{\l_tmpa_box}
  \pbs_pdfannot:nnnn{\l_box_wd_tl}{\l_box_ht_tl}{\l_box_dp_tl}{
    /Subtype/RichMedia
    /BS~<</W~0/S/S>>
    /Contents~(embedded~video~file:#3)
    /NM~(rma:#3)
    /AP~<</N~\pbs_pdflastxform:>>
    /RichMediaSettings~\pbs_pdflastobj:
    /RichMediaContent~\rmcontent
  }
  \phantom{#2}
  \group_end:
}
\ExplSyntaxOff

\usepackage{graphicx}

\newcommand{\mycomment}[1]{}

\newcommand{\vcomment}[1]{}

%% file: sec/01_introduction.tex
\begin{figure}[t]
    \centering
        \includegraphics[width=0.99\linewidth]{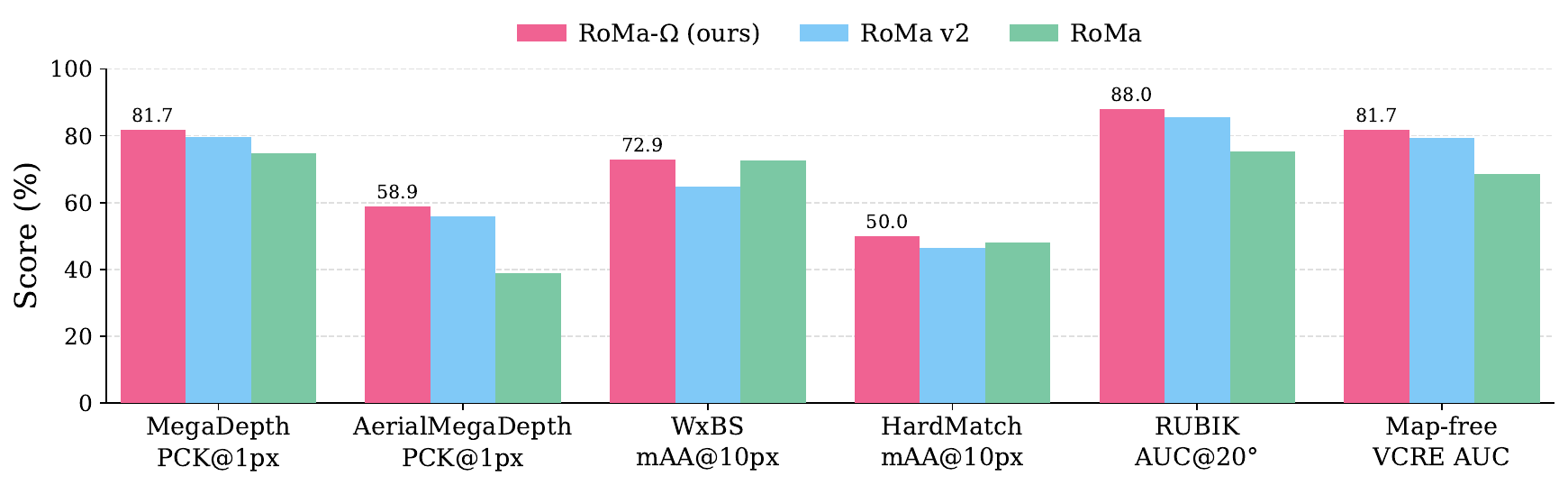}

    \caption{\textbf{\ours.} We simply replace the DINOv3 backbone of RoMa~v2 with carefully selected features from the VGGT-$\Omega$ architecture. This small change leads to our model, \ours, which achieves state-of-the-art performance on a wide range of benchmarks, surpassing the current best matchers RoMa and RoMa~v2. In particular, \ours~surpasses RoMa on the difficult matching benchmarks WxBS and HardMatch, which RoMa~v2 did not.}
    \label{fig:teaser}
\end{figure}

\section{Introduction}

Image matching is a central problem in 3D computer vision with many downstream tasks such as Structure-from-Motion (SfM)~\cite{hartley2003multiple},  Visual Localization~\cite{sattler2018benchmarking, taira2018inloc, sarlin2019coarse} and 3D Reconstruction~\cite{lee2025dense, schoenberger2016sfm, snavely2008modeling, hartley2003multiple}.  
Recently, learned dense image matchers such as RoMa~\cite{edstedt2024roma} and RoMa~v2~\cite{edstedt2026romav2} have surpassed their sparse counterparts based on keypoint detection, description~\cite{lowe2004distinctive,rublee2011orb} and matching~\cite{detone2018superpoint,sarlin2020superglue}, demonstrating exceptional generalization by replacing the traditional keypoint-based pipeline with direct per-pixel correspondence estimation.
Much of the robustness of RoMa has been attributed~\cite{edstedt2024roma, edstedt2026romav2, nordstrom2026lomalocalfeaturematching} to its use of frozen DINO features~\cite{caron2021dino, oquab2023dinov2, siméoni2025dinov3}.\footnote{As RoMa and RoMa~v2 use DINOv2 and DINOv3, respectively, we use the term DINO throughout the paper to refer to them broadly. All experiments use DINOv3.} In this paper, we show that frozen DINO features are not optimal for image matching and propose a better option.

A very recent alternative approach to SfM is to directly learn to regress 3D points and camera poses using a neural network~\cite{wang2024dust3r, wang2025vggt, wang2026vggtomega, keetha2025mapanything, depthanything3}. Such \textit{feed-forward reconstruction} approaches have gained widespread adoption. It remains unclear to what extent these models implicitly learn image matching~\cite{an2025cross, nordstrom2026mum, zhang2026emergentextremeviewgeometry3d, wang2026vggtomega}. An et al.~\cite{an2025cross} investigated the cross-attention map of CroCo~\cite{weinzaepfel2022croco} as a zero-shot matcher and Zhang et al.~\cite{zhang2026emergentextremeviewgeometry3d} studied VGGT's~\cite{wang2025vggt} attention map in the context of non-overlapping views. In this paper, we instead focus on the matching capabilities of feed-forward reconstruction models.

State-of-the-art image matchers, including RoMa and RoMa~v2~\cite{edstedt2024dedode, bökman2024steerers, edstedt2024roma, zhang2025ufm, edstedt2026romav2, nordstrom2026lomalocalfeaturematching}, do not leverage feed-forward reconstruction models for feature extraction. Instead, they typically rely on DINO features~\cite{caron2021dino, oquab2023dinov2, siméoni2025dinov3}. Nordström et al.~\cite{nordstrom2026mum} suggest that DINO features are suboptimal for image matching as they can be beaten by a simple multi-view MAE objective. Furthermore, the DINO encoder in the VGGT architecture~\cite{wang2025vggt, wang2026vggtomega} is fine-tuned to adapt to the SfM task.

We thus ask the following question: \textit{what do feed-forward 3D models know about image matching?} In particular, we study how the features compare to those of DINO. Based on these observations, it is reasonable to hope that replacing DINO features in image matchers by features from feed-forward reconstruction models would yield better performance. 

\noindent\textbf{Our investigation comprises three focus areas:}
\begin{enumerate}[label=(\roman*), topsep=0pt]
    \item \textbf{Zero-shot matching of patch features.} We systematically compare the quantitative image matching performance of features from feed-forward reconstruction models with the standard image backbone DINO~(\cref{subsec:lightweight}). We also discover a pattern of feature degradation, both qualitatively (\cref{fig:feature-corr}) and quantitatively (\cref{fig:matchbench}), in the later layers and discuss its connection to Gram anchoring in DINOv3~\cite{siméoni2025dinov3}. However, we find that the features remain highly effective when used with a linear probe.
    \item \textbf{Direct matching of 3D point predictions.} We evaluate the matching performance of the raw geometric predictions of VGGT-$\Omega$. We find this training-free baseline to be highly competitive (\cref{tab:raw-predictions}), although its performance degrades significantly upon viewpoint and modality shifts (\cref{tab:additional-evals}).
    \item \textbf{Incorporation into a full matching pipeline.} We show that state-of-the-art matching performance (\cref{tab:additional-evals,tab:visloc,tab:dense}) can be achieved by simply replacing the DINOv3 backbone of RoMa~v2 with carefully selected features from the VGGT-$\Omega$ architecture. In particular, we improve mAA by +8.1 and +3.5 on the challenging WxBS and HardMatch datasets, respectively.
\end{enumerate}

%% file: sec/02_related_work.tex
\begin{figure}[t]
    \centering
        \includegraphics[width=0.99\linewidth]{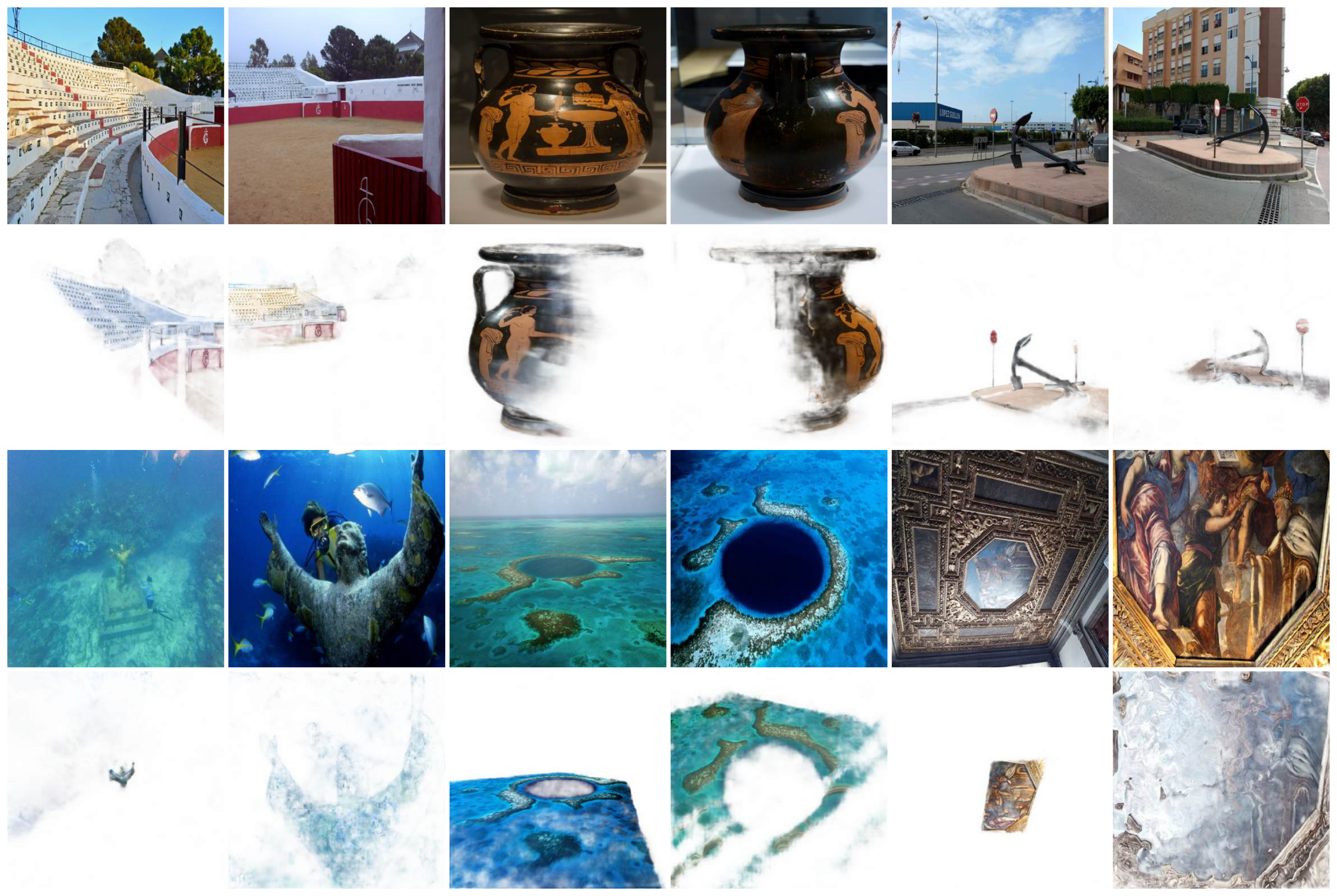}

    \caption{\textbf{Qualitative Results.} \ours~handles diverse pairs with challenging viewpoint and lighting changes. Images taken from the HardMatch~\cite{nordstrom2026lomalocalfeaturematching} dataset.}
    \label{fig:qualitative}
\end{figure}

\section{Related Work}

\subsubsection{Image Matching.}
Establishing correspondences between images is a foundational problem in 3D vision, underpinning tasks ranging from camera pose estimation to large-scale reconstruction. Classical matching pipelines decompose the problem into three stages: keypoint detection, local feature description, and correspondence assignment via nearest-neighbor search in descriptor space. Much work has been done to replace hand-crafted components such as SIFT~\cite{lowe2004distinctive} and ORB~\cite{rublee2011orb} with learned alternatives, including learned detectors~\cite{barroso2019key, mishkin2018repeatability, verdie2015tilde, edstedt2024dedodev2, edstedt2025dad}, descriptors~\cite{balntas2017hpatches, tian2019sosnet, germain2020s2dnet, edstedt2024dedode}, joint detector--descriptor architectures~\cite{detone2018superpoint, revaud2019r2d2, tyszkiewicz2020disk, Wang_2021_ICCV, zhao2022alike, Zhao2023ALIKED}, and matchers~\cite{sarlin2020superglue,lindenberger2023lightglue,nordstrom2026lomalocalfeaturematching}. Proposing to replace the whole matching pipeline, detector-free methods such as LoFTR~\cite{sun2021loftr} emerged. The field has since shifted toward dense correspondence estimation. Starting with DKM~\cite{edstedt2023dkm}, dense matchers have consistently dominated standard matching benchmarks~\cite{dai2017scannet, li2018megadepth, mishkin2015WXBS}, leading to a succession of increasingly capable methods such as RoMa~\cite{edstedt2024roma}, UFM~\cite{zhang2025ufm}, and RoMa v2~\cite{edstedt2026romav2}. More recently, image matching has become tightly integrated with feed-forward 3D reconstruction, with models such as MASt3R~\cite{leroy2024grounding} and VGGT~\cite{wang2025vggt, wang2026vggtomega} incorporating explicit correspondence objectives. In this work, we build upon this line of research by studying what feed-forward 3D models know about image matching. In particular, we show that state-of-the-art matching performance can be achieved when incorporating them as frozen feature extractors.

\subsubsection{Feed-Forward Reconstruction.}
Recovering scene geometry and camera parameters from images, commonly referred to as Structure-from-Motion (SfM)~\cite{hartley2003multiple}, has traditionally relied on multi-stage optimization pipelines such as Bundler~\cite{snavely2008modeling} and COLMAP~\cite{schoenberger2016sfm}. These systems construct sparse reconstructions by alternating between correspondence estimation, geometric verification, triangulation, and bundle adjustment, with image matching serving as the primary source of geometric constraints. Recent years have seen the emergence of feed-forward alternatives that seek to infer scene geometry directly from images using learned models. Early approaches such as DUSt3R~\cite{wang2024dust3r} and MASt3R~\cite{leroy2024grounding} directly infer 3D pointmaps from image pairs. Subsequent work extended this paradigm to larger image collections and video sequences. In particular, VGGT~\cite{wang2025vggt}  and more  recently, VGGT-$\Omega$~\cite{wang2026vggtomega}, use a large multi-view transformer to estimate all relevant 3D quantities. The pursuit of understanding these models have led to many prior works~\cite{an2025cross, stary2025understanding, burzio2026dejaview, zhang2026emergentextremeviewgeometry3d}. In contrast to prior work, we investigate their usefulness as image matchers.

%% file: sec/03_method.tex
\begin{figure}[t]
    \centering
        \includegraphics[width=0.99\linewidth]{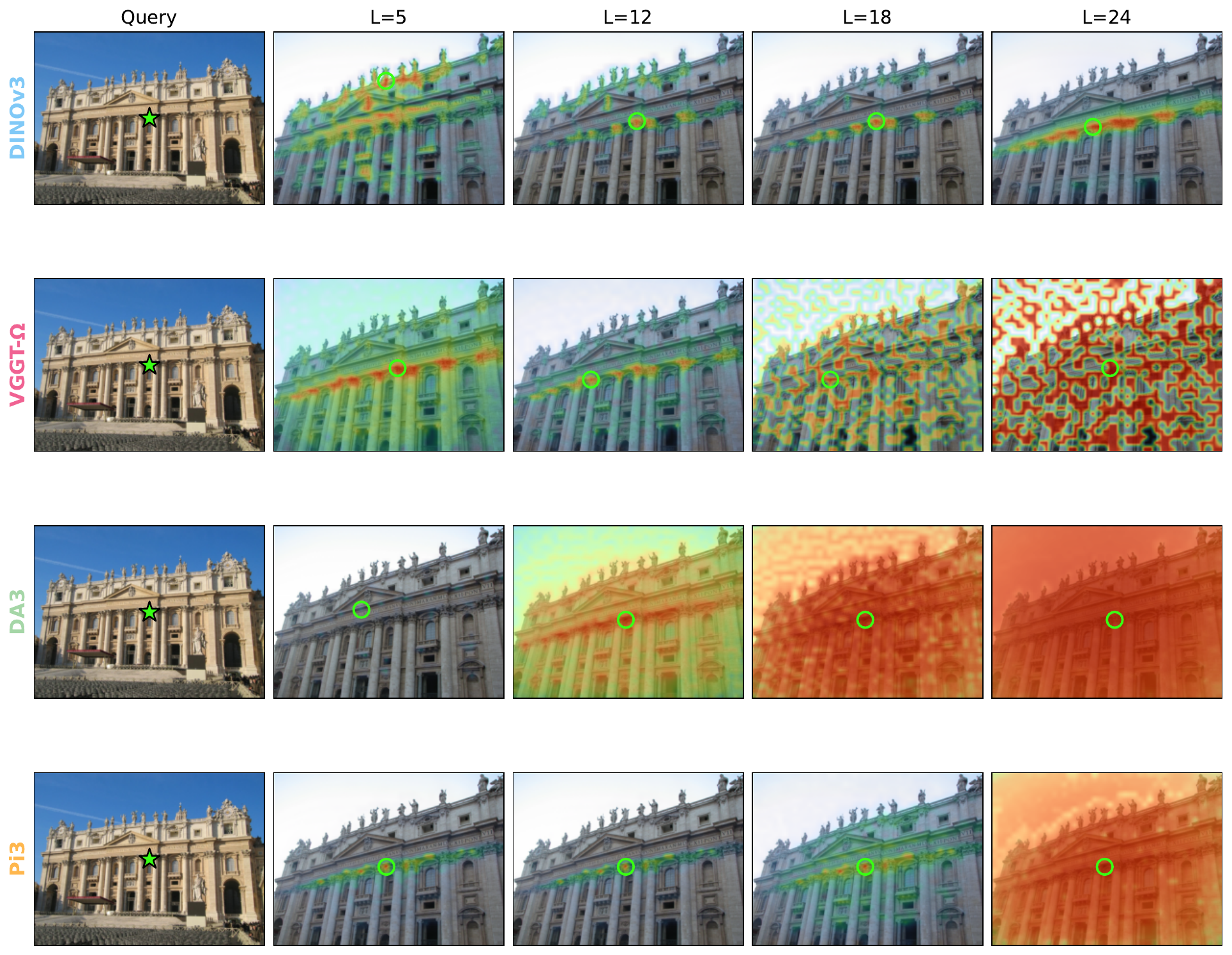}

    \caption{\textbf{Feed-forward reconstruction models, in contrast to DINO, show systematic degradation in feature correlations at later layers.} We query an image patch (green star) and plot its feature correlations in the reference image along with its maximum (green circle).}
    \label{fig:feature-corr}
\end{figure}

\section{Feed-forward Reconstruction Models for Matching}

In this section, we first provide preliminaries on feed-forward reconstruction (\cref{subsec:feedforward}) and thereafter systematically compare the frozen features of feed-forward reconstruction models to DINOv3, both qualitatively (\cref{subsec:qualitative}) and quantitatively (\cref{subsec:lightweight,subsec:full-matching}).

\subsection{Feature Extraction from Reconstruction Models}\label{subsec:feedforward}

Feed-forward reconstruction models infer scene geometry from one or more images by jointly reasoning across views. While these models are typically used to predict quantities such as point maps, depth maps, and camera poses, we instead study their internal representations as dense visual descriptors. In \cref{subsec:raw-predictions}, we also study the usefulness of the raw geometry predictions for image matching.

Formally, given an image set $\mathcal{I} = {I_i}_{i=1}^{M} \>$, the visual encoder first extracts patch features $F_i^{0}=E(I_i)\in\mathbb{R}^{P\times C} \>$, where $P$ denotes the number of image patches and $C$ the feature dimension. These features are processed by a transformer with alternating frame-wise and global attention layers, producing feature representations $F_i^{\ell}\in\mathbb{R}^{P\times C} \>$, for $\qquad
\ell=1,\dots,L$.

The final features $F_i^{L}$ are used by reconstruction heads to predict the scene geometry. In this work, we discard the reconstruction heads and evaluate the intermediate representations $F_i^\ell$ directly for image matching (except for \cref{subsec:raw-predictions}).

Unlike DINO, which is trained with self-supervised visual objectives, feed-forward reconstruction models are optimized to establish geometric consistency across views. Different layers may serve different roles: earlier layers may preserve local appearance and position, while later layers may become more task- and scene-conditioned for reconstruction. We therefore hypothesize that their hidden representations encode stronger correspondence information that can be exploited by a matching model.

\subsection{Qualitative Feature Analysis}
\label{subsec:qualitative}

\begin{figure}[t]
    \centering
        \includegraphics[width=0.99\linewidth]{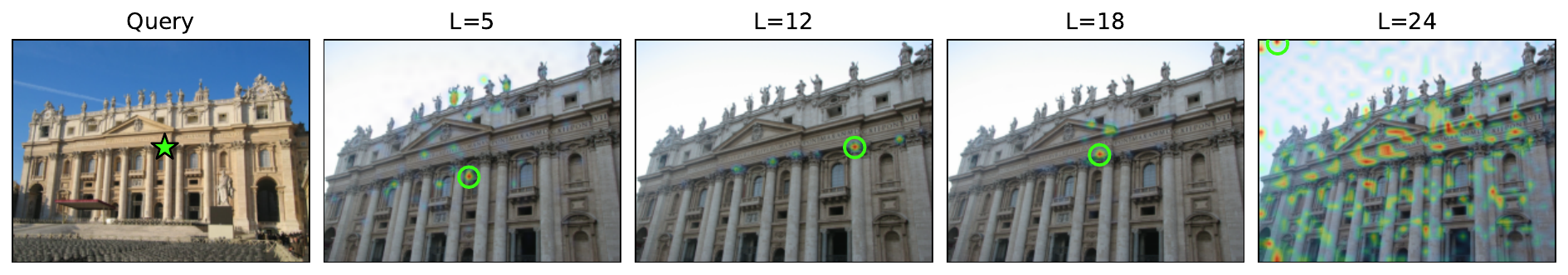}
    \caption{\textbf{VGGT-$\Omega$ attention maps encode some matching information.} We visualize the global attention map and its peak (green circle). Clearly, the attention map encodes some matching information. However, it is significantly more interpretable in the middle layers than at the last layer.}
    \label{fig:attention-map}
\end{figure}

To visualize the matching behavior of the learned features, we compute feature correlation maps between two images. Given a query location $u$ in image $I^A$ and feature vectors $f_u$ and $f_v$ from a chosen layer, we define the correlation map
\begin{equation}
C(u,v)=
\frac{f_u^\top f_v}
{\|f_u\|_2 \,\|f_v\|_2} \> ,
\end{equation}
for all locations $v$ in image $I^B$.

Figure~\ref{fig:feature-corr} compares correlation maps obtained from DINOv3 and popular feed-forward reconstruction models. As expected, DINO produces increasingly discriminative representations in later layers, resulting in localized correlation peaks near the true correspondence. In contrast, the later layers of all feed-forward reconstruction models exhibit substantially more diffuse correlation maps despite often retaining the correct correspondence as the global maximum. Similar behavior is observed in the attention maps shown in \cref{fig:attention-map}.

We hypothesize that this behavior is related to the degradation of dense features observed in DINOv3~\cite{siméoni2025dinov3}, where patch-level similarity can become unreliable even as global representations improve. The key difference is that DINOv3 explicitly addresses this issue with Gram anchoring, whereas feed-forward reconstruction models optimize for reconstruction quality and are not constrained to preserve pairwise patch similarity. These observations suggest that correspondence information remains present in the representations of feed-forward reconstruction models, but is not directly accessible through raw feature similarity. We investigate this hypothesis quantitatively in the next section.

\subsection{Lightweight Matching Experiments} 
\label{subsec:lightweight}

We evaluate the frozen representations under two settings:

\subsubsection{Zero-shot matching.}
Given a query location $u$, the predicted correspondence is obtained via nearest-neighbor retrieval in feature space:
\begin{equation}
\arg\max_{v} \ C(u,v) \> ,
\end{equation}
where $C(\cdot,\cdot)$ denotes cosine similarity.

\subsubsection{Linear probing.}
To measure the information content of the representations independently of their geometry in feature space, we train a lightweight probe on top of frozen features. Specifically, a learned linear projection maps the backbone features into a matching space, after which a small decoder predicts a dense correspondence field. Only the probe parameters are optimized.

\begin{figure}[t]
    \centering
        \includegraphics[width=0.99\linewidth]{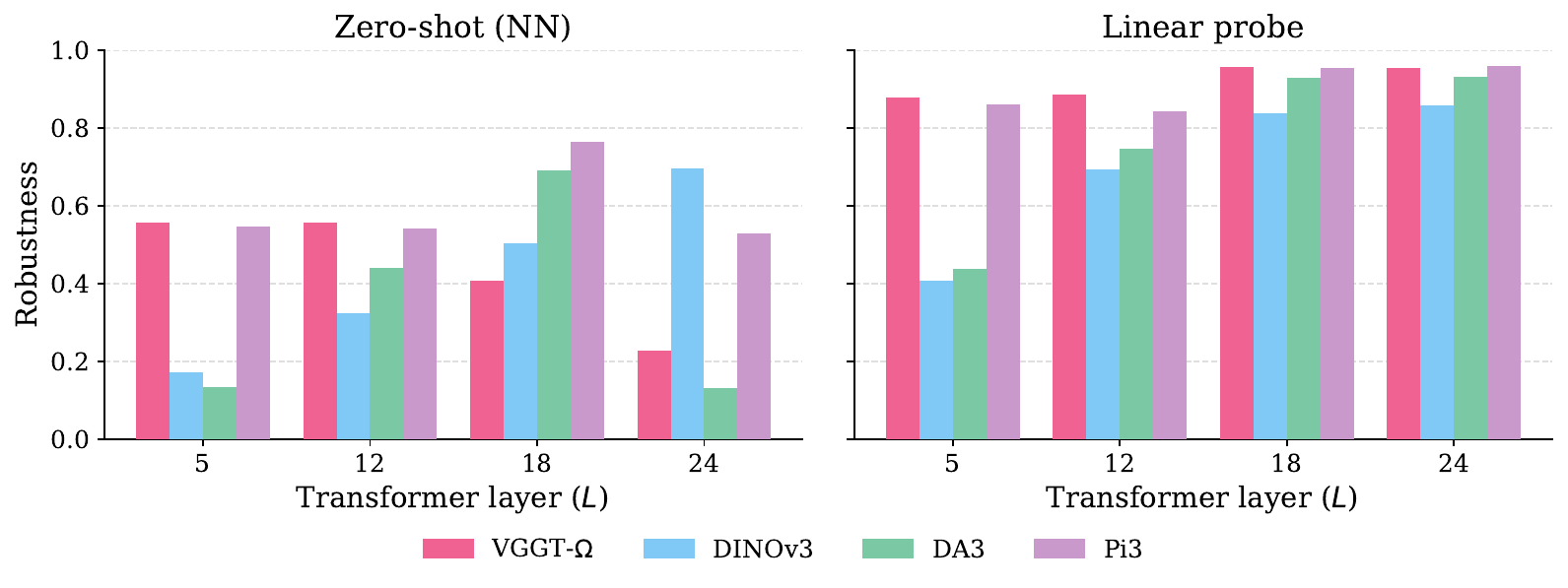}

    \caption{\textbf{Feed-forward reconstruction models require linear probing for accurate matching.} We evaluate the zero-shot (nearest-neighbor) and linear probing matching performance (PCK@32px) of features from different layers on MegaDepth-1500~\cite{li2018megadepth, sun2021loftr}. While DINO works well on its raw features, the later layers of feed-forward reconstruction models largely collapse while linear probing performance is strong.}
    \label{fig:matchbench}
\end{figure}

In \cref{fig:matchbench}, we compare features extracted from DINOv3, VGGT-$\Omega$, $\pi^3$~\cite{wang2026pi}, and DA3~\cite{depthanything3} in terms of robustness (PCK@32px) on MegaDepth-1500. While DINOv3 performs strongly under nearest-neighbor retrieval, the later layers of all reconstruction models exhibit a substantial drop in zero-shot performance. This confirms the degradation suggested by the correlation visualizations in Figure~\ref{fig:feature-corr}. However, the trend reverses under linear probing: the performance of the later reconstruction layers improves dramatically and consistently surpasses DINO. These results indicate that the representations contain strong correspondence information, but that this information is not organized in a matching-friendly feature space. Among the evaluated models, VGGT-$\Omega$ achieves the strongest linear-probe performance and is therefore used in the remainder of the paper.

\subsection{Integration into a Dense Matching Pipeline} \label{subsec:full-matching}

Linear probing evaluates whether matching information is present in the representation. Modern image matchers, however, employ substantially more expressive decoders. We therefore evaluate frozen reconstruction features within a state-of-the-art matching architecture. 

Specifically, we adopt the coarse matching stage of RoMa~v2. Given an image pair $(I^A, I^B)$, a frozen backbone extracts features $(F^A, F^B) \>$. These features are processed by a trainable transformer decoder $D_\theta(F^A,F^B) \>$, which performs cross-image reasoning. A DPT~\cite{ranftl2021vision} prediction head then converts the refined features into a dense correspondence field. During training, only the decoder and the prediction head are optimized.

\begin{figure}[t]
    \centering
        \includegraphics[width=0.99\linewidth]{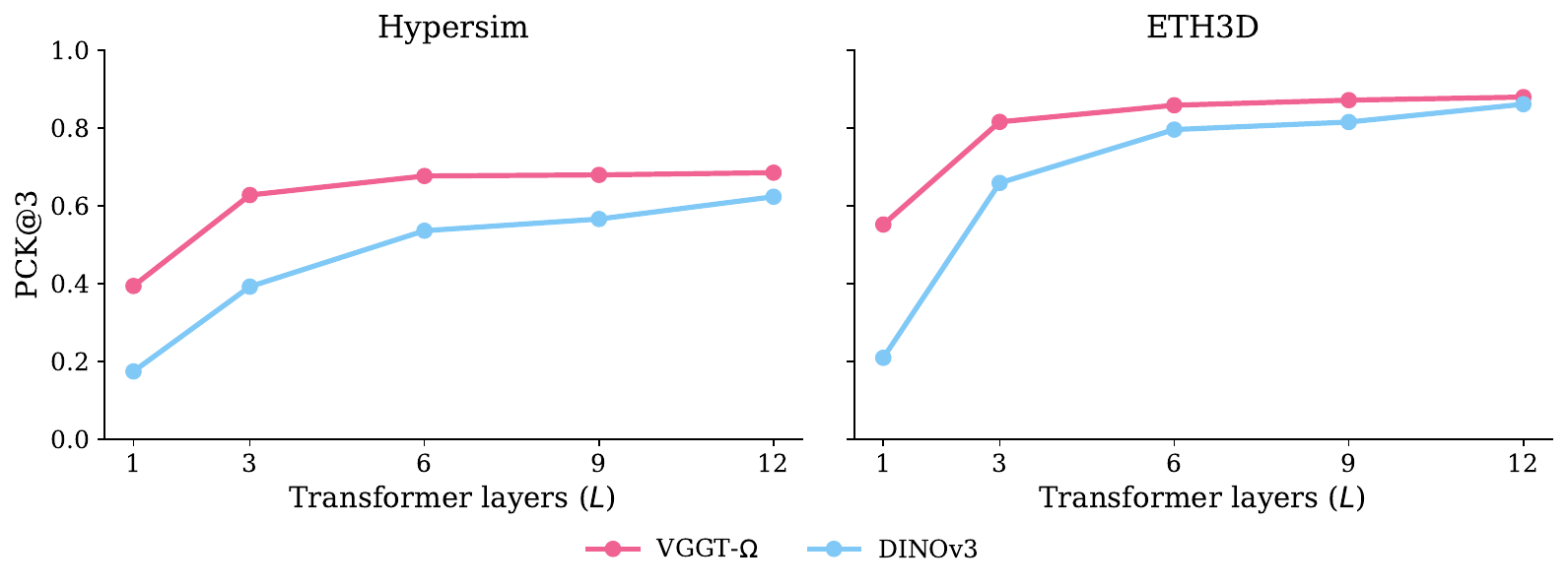}

    \caption{\textbf{Dense Matching Pipeline.} We now add $L$ transformer layers and a DPT head, mirroring RoMa~v2, and show matching performance as a function of $L$.}
    \label{fig:full-matching}
\end{figure}

We compare DINOv3 and VGGT-$\Omega$ as frozen backbones while keeping the remainder of the architecture unchanged. Figure~\ref{fig:full-matching} reports the performance as a function of the depth of the decoder. VGGT-$\Omega$ consistently outperforms DINOv3 across all decoder sizes, indicating that feed-forward reconstruction models provide a stronger foundation for dense matching. As the decoder capacity increases, the gap gradually narrows, suggesting that sufficiently powerful decoders can partially compensate for weaker backbone representations.

\section{\ours}

Given the strong performance of VGGT-$\Omega$ as a backbone for image matching demonstrated in the previous section, we now proceed with designing a matcher on top of those frozen features. To provide a clean comparison, we simply replace the frozen DINOv3 encoder in RoMa~v2, the current best image matcher, with VGGT-$\Omega$ and retrain it. In the following section, we detail a brief background on the RoMa architecture (\cref{subsec:roma}), how to extract features from VGGT~(\cref{subsec:vggt-feature-extraction}), and implementation details~(\cref{subsec:training}).

\subsection{RoMa}
\label{subsec:roma}

Dense image matching aims to establish pixel-wise correspondences between two images \(I^A\) and \(I^B\). Following RoMa, the output of the matcher consists of a dense warp $W^{A \rightarrow B} \in \mathbb{R}^{H^A \times W^A \times 2} \>$, and a confidence map $P^{A \rightarrow B} \in [0,1]^{H^A \times W^A} \>$, where \(W^{A \rightarrow B}(u)\) denotes the predicted correspondence in image \(I^B\) for pixel \(u\) in image \(I^A\), and \(P^{A \rightarrow B}(u)\) estimates the probability that the correspondence is valid (supervised by the ground-truth co-visibility).

RoMa decomposes matching into a coarse matching stage followed by a cascade of refinement modules. First, frozen visual features are extracted independently from the two images. A transformer-based match decoder then performs cross-image reasoning and predicts a coarse dense warp together with a confidence estimate. Finally, a sequence of convolutional refiners progressively increases the spatial resolution of the prediction. In RoMa, these are jointly trained, whereas in RoMa~v2 they are trained separately.

Our model follows the RoMa~v2 architecture and differs only in the visual backbone. Specifically, we replace the frozen DINO encoder with a frozen VGGT-\(\Omega\) encoder while keeping the decoder, prediction heads, losses, and training procedure unchanged. This design isolates the effect of the backbone and allows us to directly evaluate the impact of feed-forward reconstruction features on dense image matching. However, it is not obvious which VGGT-$\Omega$ features are most suitable for matching. This is the topic of our next section.

\subsection{VGGT-$\Omega$ Feature Extraction}\label{subsec:vggt-feature-extraction}

Our analysis (\cref{subsec:lightweight}) shows that intermediate VGGT-$\Omega$ representations contain strong matching information. We therefore replace the frozen DINO backbone in RoMa~v2 with frozen VGGT-$\Omega$ features while leaving the remainder of the coarse matching pipeline unchanged.

\subsubsection{Feature extraction.}
VGGT-$\Omega$ is a 24-layer transformer with alternating frame-wise and cross-view attention. We extract patch features from layers $\ell\in\{5,12,18,24\}$ and concatenate the corresponding frame-wise and cross-view token streams to obtain multi-scale feature representations.

\subsubsection{Cross-view refinement.}
Following RoMa~v2, the deepest features are refined by a transformer decoder $D_\theta$ that performs joint reasoning across both images,
\begin{equation}
\tilde{F}_i^{24}
=
F_i^{24}
+
D_\theta(F_A^{24},F_B^{24})_i,
\end{equation}
the remaining feature levels are not refined.

\subsubsection{Matching head.}
The refined deepest features are projected into the matching space and correlated as in RoMa~v2 to produce a match embedding. The DPT prediction head then consumes the two deepest feature levels ($\ell\in\{18,24\}$), together with the match embedding, to predict the coarse warp and confidence.

Aside from replacing the frozen DINO features with VGGT-$\Omega$ representations, the coarse matching pipeline is identical to RoMa~v2. We ablate the extracted feature layers in \cref{tab:ablation} and the decoder depth in \cref{fig:full-matching}.

\begin{figure}[t]
    \centering
        \includegraphics[width=0.99\linewidth]{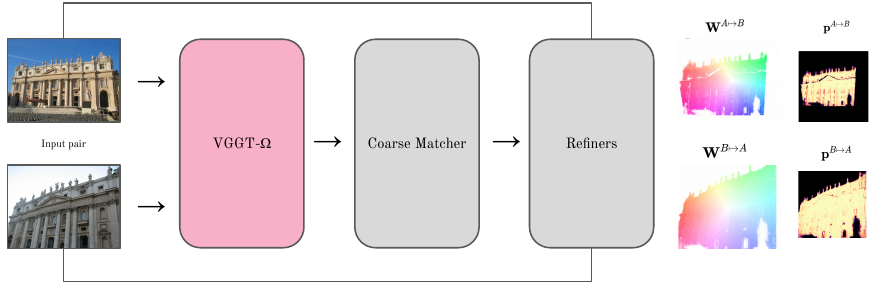}

    \caption{\textbf{\ours~architecture.} Our model, \ours, replaces DINO as the frozen visual feature extractor in RoMa~v2 with VGGT-$\Omega$ and trains a transformer decoder on top. Similar to RoMa~v2, it uses a DPT head to predict a coarse warp and refines it using convolutional refiners.}
    \label{fig:method}
\end{figure}

\subsection{Implementation Details}\label{subsec:training}

Following RoMa~v2, we train the coarse matcher on a mix of resolutions and aspect ratios, specifically: {\small \(\{512\times512,\,592\times448,\,624\times416,\,688\times384,\,448\times592,\,416\times624,\,384\times688\}\)}. The refiners are trained exclusively with size $640\times640$. We use a learning rate of $4\times10^{-4}$ for both the coarse matcher and the refiners. The coarse matcher is trained for 500K steps with a batch size of $128$, taking approximately 2 days on 16 $\times$ A100 GPUs. The refiners are trained for 300K steps with a batch size $64$, taking approximately 2 days on 8 $\times$ A100 GPUs. We train \ours~on a diverse collection of 3D vision datasets that is similar to the mix used in LoMa~\cite{nordstrom2026lomalocalfeaturematching} and RoMa~v2~\cite{edstedt2026romav2}, which is listed in the supplementary (\cref{tab:dataset_mix}).  In \cref{append:training-instabilities}, we discuss mitigating instabilities encountered during training.

%% file: sec/04_experiments.tex
\section{Experiments}
\label{sec:experiments}
In this section, we begin by studying how well VGGT-$\Omega$ matches directly from its raw predictions, without any matching-specific training (\cref{subsec:raw-predictions}). Motivated by its collapse under viewpoint changes and modality shifts, we compare \ours~to state-of-the-art image matchers on a large collection of benchmarks for extreme matching (\cref{subsec:wxbs,subsec:hardmatch,subsec:rubik}), visual localization (\cref{subsec:visloc}), and dense matching (\cref{subsec:dense}). Finally, we study the runtime (\cref{subsec:runtime}) and conduct ablations (\cref{subsec:ablations}). We primarily compare against RoMa and RoMa~v2, the current state-of-the-art dense matchers.

\subsection{Matching from Raw Predictions}\label{subsec:raw-predictions}

Before training any matcher on top of VGGT-$\Omega$, it is worth asking how well the model already matches out of the box, using only its raw geometric predictions. We extract correspondences either by warping pixels across views through the predicted per-frame depth and cameras, or by mutual nearest neighbors in the predicted 3D points, and recover relative pose with an essential matrix solver and RANSAC. As a reference, we also read the relative pose directly from the two predicted cameras. As reported in \cref{tab:raw-predictions} on ScanNet-1500~\cite{dai2017scannet,sarlin2020superglue}, this training-free baseline is already competitive: depth warping is the strongest correspondence extractor and essentially matches the pose read directly from the cameras, while mutual nearest neighbors in 3D lag behind. Perhaps surprisingly, both slightly exceed RoMa~v2 and the original VGGT tracking pipeline~\cite{wang2025vggt}. This strength is, however, confined to moderate viewpoint and domain shifts.

\begin{table}[t]
\centering
\small
\caption{\textbf{Matching from VGGT-$\Omega$ raw predictions on ScanNet-1500}. Relative-pose AUC recovered from correspondences extracted from VGGT-$\Omega$'s raw predictions, by depth warping or mutual nearest neighbors in 3D, and, as a reference, from the predicted cameras directly (\emph{Direct}). Without any matching-specific training, VGGT-$\Omega$ is on par with or slightly above RoMa~v2 and the original VGGT tracking pipeline.}
\label{tab:raw-predictions}
\begin{tabular}{l rrr}
\toprule
Method & AUC@$5^\circ$ & AUC@$10^\circ$ & AUC@$20^\circ$ \\
\midrule
\multicolumn{4}{@{}l@{}}{\small \textit{VGGT-$\Omega$ raw predictions}} \\
Depth warp & \bfseries 34.2 & \bfseries 57.6 & 74.8 \\
Mutual NN in 3D & 22.7 & 43.0 & 59.5 \\
Direct & \bfseries 34.2 & \bfseries 57.6 & \bfseries 74.9 \\
\midrule
VGGT~\cite{wang2025vggt} & 33.9 & 55.2 & 73.4 \\
\rowcolor{romav2row} RoMa~v2~\cite{edstedt2026romav2} & 33.6 & 56.2 & 73.8 \\
\bottomrule
\end{tabular}
\end{table} 

As the VGGT-$\Omega$ rows of \cref{tab:additional-evals} show, the raw predictions stay strong on the geometric challenges of RUBIK ($84.3$ AUC@20$^\circ$) but collapse under the extreme illumination and modality changes of WxBS and HardMatch ($29.0$ and $20.2$ mAA). We attribute this to the training regime of feed-forward reconstruction models, which sample nearby frames from videos and are therefore mostly exposed to temporally close, easy-to-match views. This motivates learning a dedicated matcher on top of the frozen features, \ie, \ours, rather than relying on the raw predictions.

\subsection{Multi-Modal Matching on WxBS}\label{subsec:wxbs}

We evaluate the robustness of \ours~on the extremely challenging WxBS benchmark~\cite{mishkin2015WXBS}. 
This benchmark consists of hand-labeled correspondences between images taken with extreme changes in either viewpoint, illumination, modality, or all three, making it a challenging test of out-of-distribution generalization..
Results are presented in~\Cref{tab:additional-evals}. \ours~beats RoMa~v2 and RoMa by +8.1 mAA and +0.3, respectively. \ours~achieves results that combine the robust generalization of RoMa and the sub-pixel accuracy of RoMa~v2.

\subsection{Diverse Matching on HardMatch}\label{subsec:hardmatch}

HardMatch is a new matching benchmark released by Nordström et al.~\cite{nordstrom2026lomalocalfeaturematching}. The dataset features 1,000 pairs from 100 different categories. The pairs vary in difficulty with many being extremely difficult. We highlight some qualitative examples in \cref{fig:qualitative}. We report the results for \ours~in \cref{tab:additional-evals}. We find that \ours~is more robust than RoMa~v2 and improves its score by +3.5 mAA.

\subsection{Matching across Geometric Challenges on RUBIK}\label{subsec:rubik}

We evaluate \ours~on RUBIK~\cite{loiseau2025rubik}, a subset of NuScenes~\cite{caesar2020nuscenes}, and report the AUC@20$^\circ$ in \cref{tab:additional-evals}. \ours~provides a meaningful improvement over the previous state-of-the-art RoMa~v2 by +2.4 AUC@20$^\circ$.

\begin{table}[t]
\small
\centering
\caption{\textbf{Comparison on the WxBS~\cite{mishkin2015WXBS}, HardMatch~\cite{nordstrom2026lomalocalfeaturematching}, and RUBIK~\cite{loiseau2025rubik}.}} 

\begin{tabular}{l r r r}
  \toprule
        Method & $\>$ WxBS (mAA$@$10px)& $\>$ HardMatch (mAA$@$10px)& $\>$ RUBIK (AUC@20$^\circ$)\\
 \midrule
VGGT-$\Omega$ & 29.0 & 20.2 & 84.3 \\
\rowcolor{romarow} RoMa & 72.6 & 48.1 & 75.4 \\
\rowcolor{romav2row} RoMa~v2 & 64.8 & 46.5 & 85.6 \\
\rowcolor{oursrow} \ours & \bfseries 72.9 & \bfseries 50.0 & \bfseries 88.0\\
  \bottomrule
\end{tabular}
\label{tab:additional-evals}
\end{table}

\subsection{Visual Localization on Map-free}\label{subsec:visloc}

The map-free relocalization benchmark~\cite{arnold2022map} tests the ability to localize the camera in metric space given a single reference image and no map. 
To obtain monocular metric depth, we use DA3~\cite{depthanything3}. We submit our results to the official online benchmark and report the results for the set in \cref{tab:visloc}.
\ours~achieves a +2.3 improvement in AUC VCRE (<45px) compared to RoMa~v2.

\begin{table}
\centering
\caption{\textbf{Visual localization.} Comparison on the test set of Map-free~\cite{arnold2022map}.}
\label{tab:visloc}
\small

\begin{tabular}{l r rr r rr}
\toprule
Method
&& \multicolumn{2}{c}{VCRE (<90px)}
&$\quad$& \multicolumn{2}{c}{VCRE (<45px)} \\
\cmidrule(lr){3-4} \cmidrule(lr){6-7} 

&& Precision~$\uparrow$ & \hspace{0.1cm} AUC~$\uparrow$
&& Precision~$\uparrow$ & \hspace{0.1cm} AUC~$\uparrow$ \\
\midrule
VGGT-$\Omega$ && 73.5 & 83.9  && 54.8 & 73.0\\
\rowcolor{romarow} RoMa && 59.7 & 84.4 && 47.5 & 68.5 \\
\rowcolor{romav2row} RoMa~v2 && 78.9 & 93.3 && 59.4 & 79.4 \\
\rowcolor{oursrow} \ours && \bfseries 79.2 & \bfseries 94.5 && \bfseries 60.2 & \bfseries 81.7 \\
\bottomrule
\end{tabular}
\end{table}

\subsection{Dense Matching}\label{subsec:dense}

We evaluate dense matching performance on a wide range of datasets (\cref{tab:dense}). \ours~ outperforms RoMa~v2 in 5/6 datasets. The only dataset on which \ours~underperforms RoMa~v2 is FlyingThings3D, which features dynamic scenes.

\begin{table}
    \centering
    \caption{\textbf{Dense matching performance.}
    Images are resized to $640\times 640$.}
    \resizebox{\linewidth}{!}{%
    \begin{tabular}{l rrrr | rrrr | rrrr}
    \toprule
        \multirow{2}{*}{Method}& \multicolumn{4}{ c }{TA-WB~\cite{wang2020tartanair, zhang2025ufm}} & \multicolumn{4}{ c}{MegaDepth~\cite{li2018megadepth}} & \multicolumn{4}{ c }{ScanNet++ v2~\cite{yeshwanth2023scannet++}} \\
        \cmidrule(lr){2-5} \cmidrule(lr){6-9} \cmidrule(lr){10-13}
          & EPE~$\downarrow$ & 1px~$\uparrow$  & 3px~$\uparrow$  & 5px~$\uparrow$  & EPE~$\downarrow$ & 1px~$\uparrow$  & 3px~$\uparrow$  & 5px~$\uparrow$  & EPE~$\downarrow$ & 1px~$\uparrow$  & 3px~$\uparrow$  & 5px~$\uparrow$   \\
         \midrule
     \rowcolor{romarow} RoMa & 60.61 & 35.1 & 52.6 & 56.2 & 2.34 & 74.8 & 93.7 & 96.4& 27.52 & 20.2 & 42.8 & 53.6 \\
     \rowcolor{romav2row} RoMa~v2  & 13.82 & 67.7 & 81.8 & 85.8 & 1.47 & 79.6 & 94.7 & 96.7 & 4.00 & 45.5 & 77.3 & 86.6 \\
     \rowcolor{oursrow} \ours & \bfseries 9.24 & \bfseries 67.9 & \bfseries 84.1 & \bfseries 87.8 & \bfseries 0.97 & \bfseries 81.7 & \bfseries 96.5 & \bfseries 98.3 & \bfseries 3.85 & \bfseries 45.6 & \bfseries 77.8 & \bfseries 87.4  \\

    \midrule
     & \multicolumn{4}{ c }{FlyingThings3D~\cite{flyingthings3d}} & \multicolumn{4}{ c}{AerialMegaDepth~\cite{vuong2025aerialmegadepth}} & \multicolumn{4}{ c }{Map-free~\cite{arnold2022map}} \\
        \cmidrule(lr){2-5} \cmidrule(lr){6-9} \cmidrule(lr){10-13}

        \rowcolor{romarow} RoMa & 5.68 & 78.0 & 86.6 & 89.2 & 25.05 & 39.0 & 65.0 & 73.9& 8.55 & 45.8 & 72.3 & 80.9 \\
        \rowcolor{romav2row} RoMa~v2  & \bfseries 0.93 &  \bfseries 89.4 & \bfseries 95.2 & \bfseries 96.8 &  4.12 & 55.9 & 81.1 &  87.9& 2.03 & \bfseries 55.4 &  84.9 & 92.7 \\
     \rowcolor{oursrow} \ours & 1.07 & 88.5 & 94.6 & 96.3 & \bfseries 2.61 & \bfseries 58.9 & \bfseries 85.3 & \bfseries 91.8 & \bfseries 2.02 & \bfseries 55.4 & \bfseries 85.1 & \bfseries 93.0 \\
         \bottomrule
    \end{tabular}%
    }

    \label{tab:dense}
\end{table}

\subsection{Runtime Comparisons}\label{subsec:runtime}

Replacing the DINO backbone used in RoMa and RoMa~v2 ($\approx300$M parameters) for VGGT-$\Omega$ ($\approx1$B parameters) naturally makes the model significantly heavier. In \cref{tab:runtime}, we benchmark the computational complexity and find that \ours~is around 40\% slower than RoMa~v2 and on par with RoMa. However, VRAM usage is significantly higher.

\begin{table}
    \centering
    \small
    \caption{\textbf{Computational complexity.} Benchmarking throughput and peak memory on $560\times 560$ images with a batch size of 8 on an A100.}
    \begin{tabular}{l rrr}
        \toprule
        Method & $\>$ Throughput (pairs/s) $\uparrow$ & $\>$ Memory (GB) $\downarrow$ & $\>$ Params. (trainable / total) $\downarrow$\\
        \midrule
            \rowcolor{romarow} RoMa & 12.7 & 3.9 &  \bfseries 111M / 414M \\
           \rowcolor{romav2row} RoMa~v2 & \bfseries 21.6 & \bfseries 4.0 &  122M / 425M \\
          \rowcolor{oursrow} \ours & 12.1 & 29.9 & 125M / 1033M \\
          \bottomrule
    \end{tabular}
    \label{tab:runtime}
\end{table}

\subsection{Ablation Studies}\label{subsec:ablations}

In \cref{tab:ablation}, we study the impact of design choices on the coarse matching performance. We begin by simply retraining RoMa~v2 on our data (\RN{1}). We then replace the DINOv3 backbone with VGGT-$\Omega$ and initially train only a DPT head on top (\RN{2}). However, this proves insufficient for strong matching performance. We reintroduce the transformer decoder from RoMa~v2 and see a substantial improvement in performance (\RN{3}). After carefully selecting the feature layers, we find that only using layers $L=\{18, 24\}$ provides the best performance (\RN{4}). Finally, we remove the feature down-projection before the DPT head (\RN{5}), allowing it to operate at the full feature dimensionality. This leads to slightly worse performance. We show full training runs in \cref{fig:training-dynamics} in the supplementary.

\begin{table}[t]
\centering
\caption{\textbf{Architecture.} Performance of the coarse matcher after 100K steps.}
\label{tab:ablation}
\small

\begin{tabular}{l rrr rrr}
\toprule
Method
& \multicolumn{3}{c}{Hypersim}
& \multicolumn{3}{c}{ETH3D} \\
\cmidrule(lr){2-4} 
\cmidrule(lr){5-7}   
PCK@$\rightarrow$ & 1px & 3px & 5px & 1px & 3px & 5px \\
\midrule

\rowcolor{gray!25} \RN{1}: RoMa~v2 (Baseline) & 21.0 & 61.8 & 74.2 & 40.6 & 85.8 & 94.2 \\
\RN{2}: VGGT-$\Omega$ backbone + DPT head & 4.9 & 28.3 & 48.0 & 8.7 & 47.5 & 70.7 \\
\RN{3}: \RN{2} + transformer decoder & 21.5 & 64.6 & 78.8 & 34.1 & 82.3 & 93.0 \\
\rowcolor{green!25} \RN{4}: \RN{3} + only 2 deepest feature levels (\ours) & \bfseries 27.8 & \bfseries 68.5 & \bfseries 80.9 & \bfseries 43.4 & \bfseries 87.7 & \bfseries 95.1 \\
\RN{5}: \RN{4} + Full width in DPT (2048) & 23.2 & 65.1 & 79.1 & 37.6 & 85.3 & 94.4 \\

\bottomrule
\end{tabular}
\end{table}

%% file: sec/05_limitations.tex
\section{Limitations}
\begin{itemize}
    \item While our investigation aims to study feed-forward 3D models broadly, we limit our largest experiments to VGGT-$\Omega$, which is among the most recent and capable feed-forward 3D models. 
    \item We pose the question: \textit{what do feed-forward 3D models know about image matching?} A natural extension is to investigate the inverse relationship: \textit{what do image matchers know about feed-forward reconstruction?} While this question is outside the scope of this paper, we believe it represents an interesting direction for future work.
    \item The practical usefulness of our model, \ours, is limited by its  computational cost, although it achieves higher accuracy than previous methods.
\end{itemize}

%% file: sec/06_conclusion.tex
\section{Conclusion}
We investigated what feed-forward 3D models know about image matching by comparing their feature representations with those of DINOv3 for zero-shot matching and when integrating them into a modern matching pipeline. Interestingly, we found that raw feature correlations degrade in the later layers for all considered feed-forward reconstruction models, and we discuss their relationship with the feature degradation observed in DINOv3. Furthermore, we studied the matching performance of VGGT-$\Omega$'s raw geometry predictions and found them to be competitive with state-of-the-art matchers. However, performance significantly degrades under viewpoint and domain shifts. Lastly, we replaced the DINO backbone in RoMa v2 with VGGT-$\Omega$, introducing \ours. Through extensive experimentation, we show that, despite being computationally expensive, \ours~outperforms RoMa and RoMa v2 across a wide range of benchmarks.

%% file: sec/acknowledgements.tex
\section*{Acknowledgements}
This work was supported by the Wallenberg Artificial
Intelligence, Autonomous Systems and Software Program
(WASP), funded by the Knut and Alice Wallenberg Foundation.
The computational resources were provided by the
National Academic Infrastructure for Supercomputing in
Sweden (NAISS) at C3SE, partially funded by the Swedish Research
Council through grant agreement no.~2022-06725, and by
the Berzelius resource, provided by the Knut and Alice Wallenberg Foundation at the National Supercomputer Centre.

%% file: sec/supplementary.tex
\clearpage
\setcounter{page}{1}
\setcounter{section}{0}

\renewcommand{\thesection}{\Alph{section}}

\begin{center}
{\Large\bf \ours: What Feed-Forward 3D Models Know About Image Matching}\\[0.5em]
{\large Supplementary Material}
\end{center}

\section{Training Instabilities}\label{append:training-instabilities}

\begin{figure}[t]
    \centering
    \begin{subfigure}[t]{0.45\linewidth}
        \centering
        \includegraphics[width=\linewidth]{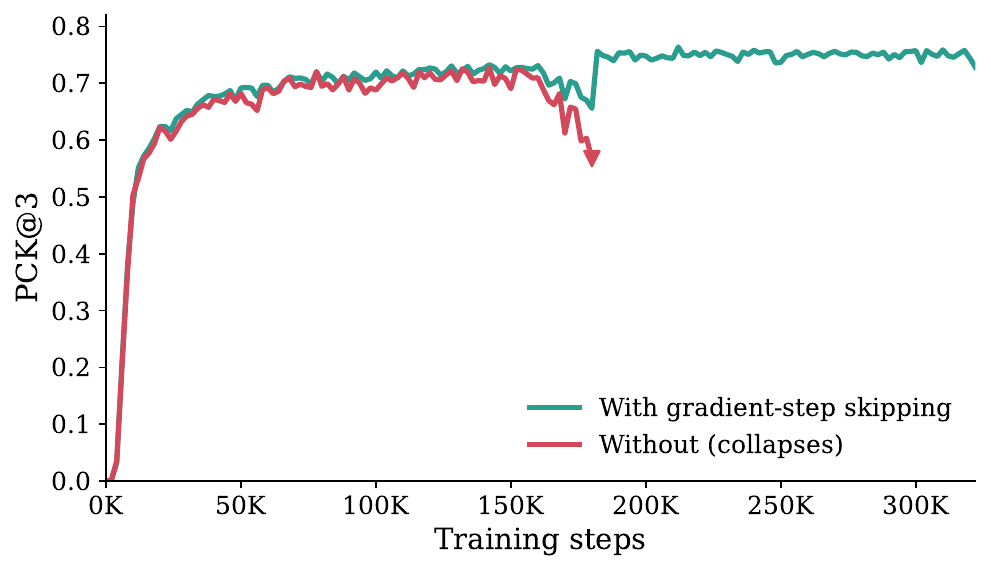}
    \end{subfigure}
    \hfill
    \begin{subfigure}[t]{0.45\linewidth}
        \centering
        \includegraphics[width=\linewidth]{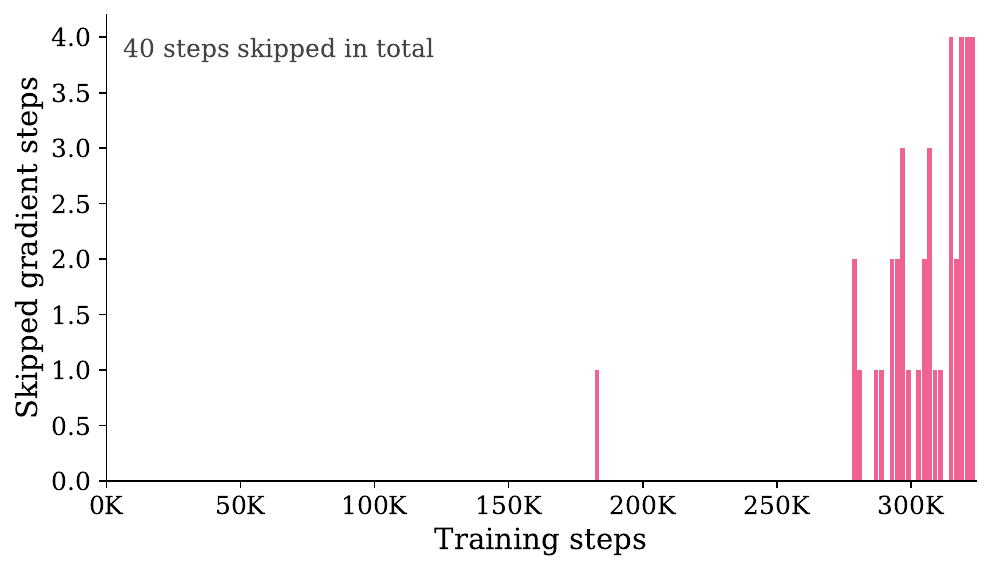}
    \end{subfigure}
    \caption{\textbf{Unstable training remedied by gradient skips.} We found that as training progresses the effect of outlier data grows. To protect against bad gradients we (i) skip gradients that are unusually high and (ii) gradually decrease the learning rate.}
    \label{fig:training-stability}
\end{figure}

When re-training RoMa v2 with a VGGT-$\Omega$ backbone, we found that training was prone to collapse. After around 150K steps the gradient norms would blow up and loss would start increasing. Not knowing if this was due to dirty data or some aspect of the training recipe, we took multiple precautions that subsequently stabilized training. The main changes were two-fold: (i) tracking EMA of gradients and rejecting gradient updates that were more than 3 times larger than the EMA and (ii) introducing a light learning-rate decay protocol. In \cref{fig:training-stability}, we illustrate its stabilizing effects.

\section{Training Dynamics}

\begin{figure}[t]
    \centering
        \includegraphics[width=0.99\linewidth]{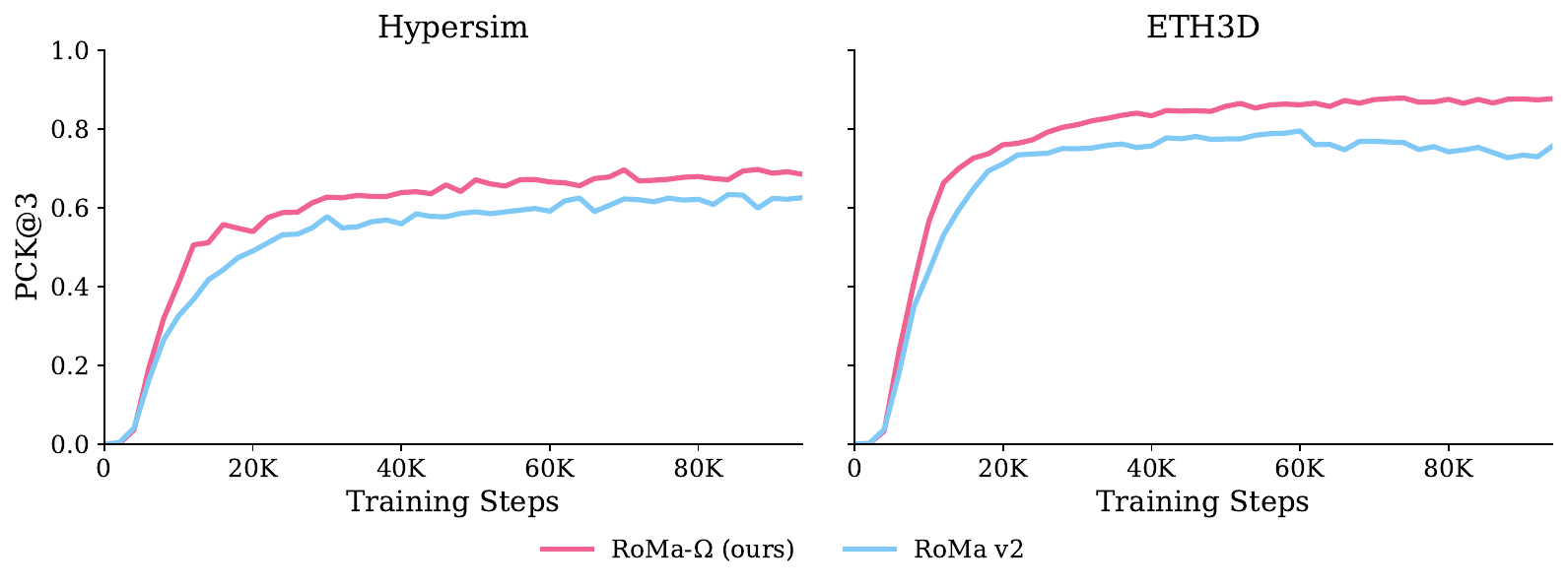}

    \caption{\textbf{Training Dynamics.} We retrain RoMa v2 on our data and compare the DINOv3 backbone to VGGT-$\Omega$ used in \ours.}
    \label{fig:training-dynamics}
\end{figure}

For fair comparison, we retrain RoMa~v2 on our data. In \cref{fig:training-dynamics}, we compare the evaluation performance during training of the two models. Clear gains are observed from using the VGGT-$\Omega$ backbone instead of the DINOv3 backbone used in RoMa v2. 

\section{Training Data}

In \cref{tab:dataset_mix}, we detail the mix of datasets used to train or model and which is the basis for all of ours experiments. This mix is inspired by those of RoMa~v2 and LoMa. While our dataset mix is slightly bigger than that of RoMa v2, we showed the prowess of our method in a fair comparison in the previous section as well as in the ablations of the main paper.

\begin{table}[htbp] \centering
\scriptsize
\caption{\textbf{Training data}. We train on a diverse mix of 3D datasets comparable to the size used for RoMa v2~\cite{edstedt2026romav2} and LoMa~\cite{nordstrom2026lomalocalfeaturematching}.}
\label{tab:dataset_mix}
\setlength{\tabcolsep}{2pt}
\begin{tabular}{lcc}
\hline
\toprule
Datasets                          & Type / GT Source            & Weight \\ 
\midrule
ScanNet++ v2~\cite{yeshwanth2023scannet++} & Indoor / Mesh & 1 \\
BlendedMVS~\cite{yao2020blendedmvs}  & Aerial / Mesh      &1 \\
Map-Free~\cite{arnold2022map}& Object-centric / MVS &1 \\
Hypersim~\cite{roberts2021hypersim}& Indoor / Graphics          &1 \\
MegaScenes~\cite{tung2024megascenes} & Outdoor / MVS & 1 \\
MegaDepth~\cite{li2018megadepth}    & Outdoor / MVS        &1 \\
MegaDepth (Re-MVS) & Outdoor / MVS        &1 \\
MegaDepth-X~\cite{li2026longtail} & Outdoor / MVS        &1 \\
AerialMD~\cite{vuong2025aerialmegadepth}    & Aerial / MVS        &1 \\
TartanAir v2~\cite{wang2020tartanair}& Outdoor / Graphics          &1 \\

Mapillary Planet-scale Depth~\cite{mpsd:2020} & Driving / MVS  & 0.1 \\
Aria Synthetic Environments~\cite{AriaSynthEnv:2025} & Indoor / Graphics & 0.1\\
CO3Dv2~\cite{co3d:2021} & Object-centric / MVS & 0.1 \\
MegaSynth~\cite{Jiang_2025_CVPR} & Indoor / Graphics & 0.1\\
SpatialVID~\cite{wang2025spatialvid} & Forward-motion / MVS & 0.01\\
\midrule
FlyingThings3D~\cite{mayer2016large} & Outdoor / Graphics & 0.5\phantom{0}\\
UnrealStereo4k~\cite{tosi2021unrealstereo4k} & Outdoor / Graphics & 0.01\\
Virtual KITTI 2~\cite{gaidon2016virtual,cabon2020vkitti2} & Outdoor / Graphics & 0.01\\

\bottomrule
\end{tabular}
\vspace{-0.5em}
\normalsize
\end{table}